\documentclass[10pt,journal,compsoc]{IEEEtran}
\usepackage{caption}
\usepackage{makecell}
\usepackage{booktabs}
\usepackage{subfig}
\usepackage{multirow}
\usepackage{graphicx}

\usepackage{url}

\usepackage{breakurl}
\usepackage[breaklinks]{hyperref}
\usepackage{enumerate}

\ifCLASSOPTIONcompsoc
  \usepackage[nocompress]{cite}
\else
  \usepackage{cite}
\fi
\ifCLASSINFOpdf
\else
\fi
\usepackage{amsmath}
\begin{document}
%
\title{Unsupervised Enhancement of Soft-biometric Privacy with Negative Face Recognition}

\author{Philipp~Terh\"{o}rst,
        Marco~Huber,
        Naser~Damer,~\IEEEmembership{Member,~IEEE,}
        Florian Kirchbucher,
        and~Arjan~Kuijper,~\IEEEmembership{Member,~IEEE}
\IEEEcompsocitemizethanks{\IEEEcompsocthanksitem The authors are with the Fraunhofer Institute for Computer Graphics Research IGD, Darmstadt, Germany, and with the Technical University of Darmstadt, Darmstadt, Germany.\protect\\
E-mail: \{forename.lastname\}@igd.fraunhofer.de
\IEEEcompsocthanksitem This work was supported by the German Federal Ministry of Education and Research (BMBF) as well as by the Hessen State Ministry for Higher Education, Research and the Arts (HMWK) within the National Research Center for Applied Cybersecurity (ATHENE), and in part by the German Federal Ministry of Education and Research (BMBF) through the Software Campus project.
Portions of the research in this paper use the FERET database of facial images collected under the FERET program, sponsored by the DOD Counterdrug Technology Development Program Office.}
}

\markboth{Journal of \LaTeX\ Class Files,~Vol.~14, No.~8, August~2015}%
{Shell \MakeLowercase{\textit{et al.}}: Bare Demo of IEEEtran.cls for Biometrics Council Journals}
%



\IEEEtitleabstractindextext{%
\begin{abstract}
Current research on soft-biometrics showed that privacy-sensitive information can be deduced from biometric templates of an individual.
Since for many applications, these templates are expected to be used for recognition purposes only, this raises major privacy issues.
Previous works focused on supervised privacy-enhancing solutions that require privacy-sensitive information about individuals and limit their application to the suppression of single and pre-defined attributes.
Consequently, they do not take into account attributes that are not considered in the training.
In this work, we present Negative Face Recognition (NFR), a novel face recognition approach that enhances the soft-biometric privacy on the template-level by representing face templates in a complementary (negative) domain.
While ordinary templates characterize facial properties of an individual, negative templates describe facial properties that does not exist for this individual.
This suppresses privacy-sensitive information from stored templates.
Experiments are conducted on two publicly available datasets captured under controlled and uncontrolled scenarios on three privacy-sensitive attributes.
The experiments demonstrate that our proposed approach reaches higher suppression rates than previous work, while maintaining higher recognition performances as well.
Unlike previous works, our approach does not require privacy-sensitive labels and offers a more comprehensive privacy-protection not limited to pre-defined attributes.

\end{abstract}

}

\maketitle

\IEEEdisplaynontitleabstractindextext

%
\IEEEpeerreviewmaketitle

\IEEEraisesectionheading{\section{Introduction}\label{sec:introduction}}

%
%
%
%

\IEEEPARstart{T}{he} face is one of the most used biometric modalities \cite{CrazyFaces}\cite{Wang2018}, because it is known as a unique biometric modality and it does not require an active user-participation \cite{DetectMorphing}\cite{Tripathi2017}.
A typical face recognition system contains feature representations (templates) for each individual enrolled.
Comparing two templates allows to verify a claimed identity or to identify an unknown subject \cite{Patil2016}.
However, recent work showed that more information than just the person's identity can be deduced from these templates \cite{7273870}. 
With the use of soft-biometric estimators, information about gender, age, ethnicity, sexual orientation or the health status can be obtained \cite{7273870}\cite{sexualOrientation}.
Since many applications do not permit to have access to these information, this shows a major invasion of privacy.
In many systems, the stored data should be exclusively used for recognition purposes \cite{8272743} and extracting such information without a person's consent can be considered as a violation of their privacy \cite{Kindt2013}.
To prevent such function creep, privacy-sensitive information should be either hidden or suppressed in face templates. 

Previous works proposed privacy-enhancing solutions for this problem that are either (a) limited to the suppression of single pre-defined attributes \cite{SemiAdNetworks}\cite{8272743}\cite{SuppressingGender} or (b) only capable of a restricted suppression performance \cite{Terhoerst2019AI}.
While (b) only offers a restricted privacy-protection, solutions from (a) are vulnerable to unconsidered function creep attacks.

In this work, we propose NFR, a novel unsupervised face recognition approach that performs comparisons of face templates in a complementary (negative) fashion.
While ordinary positive templates describes individuals how they actually are, negative templates stores only random complementary information about the individual.
This suppresses privacy-sensitive information in the template and thus, prevents function creep attackers from easily extracting these information.
In order to forecast and guarantee a certain recognition performance, we provide a theoretical reasoning of our solution and further demonstrate its correctness empirically.

Soft-biometric privacy is challenged by maintaining a high recognition performance while achieving a high suppression performance for privacy-sensitive attributes. 
Therefore, we analyse both aspects on two publicly available databases under controlled and uncontrolled circumstances.
The evaluation of the attribute suppression performance is done on three soft-biometric attributes: gender, age, and race.
Unlike most of previous works, we design our experiments in the context of a function creep attacker who knows and adapts to the used privacy mechanism.

The experiments show that our proposed approach is able to reach 2-4 times higher suppression rates than previous works under different attack mechanisms and attributes, while maintaining significantly higher recognition performances.
In the uncontrolled scenario, our solution fully retains the recognition performance while reaching suppression rates of up to 36\%.

The main contribution of this work is a privacy-enhancing solution that
\begin{enumerate}[i)]
\item works on the biometric template-level, since most biometric representations are stored in templates rather than images \cite{DBLP:conf/ncc/DeyBBDCPS14}\cite{Stokkenes:2016:BAP:2947626.2951962};
\item considers a more challenging attack scenario of an attacker that adapts to the systems privacy mechanism;
\item does not require privacy-sensitive labels about individuals (unsupervised privacy-preservation); and thus,
\item offers a more comprehensive privacy-protection which is not limited to the suppression of pre-defined attributes.
\end{enumerate}
Additionally, the source code for our approach is available at the following link\footnote{\url{https://github.com/pterhoer/PrivacyPreservingFaceRecognition/tree/master/unsupervised/negative_face_recognition}}.

\section{Related work}
\label{sec:RelatedWork}



Recent years exposed a growing interest in privacy preservation in biometrics and related fields.
In the field of data mining, a lot of work was done learning statistical database properties while ensuring individual privacy \cite{differentialPrivacy}. 
This lead to privacy-enhancing concepts such as $k$-anonymity \cite{kAnonymity}, $l$-diversity \cite{lDiversity}, $t$-close\-ness \cite{tCloseness}, and differential privacy \cite{differentialPrivacy}.

In the context of biometrics, privacy enhancing technologies aim at assuring that biometric data collected from an individual is only used for the specific purpose of the application \cite{SuppressingGender}.
For face representations, privacy preservation has been studied from two perspectives.
Either it focuses on preserving facial characteristics like gender, age, and expression while de-identifying face images \cite{1640608}\cite{7139096}\cite{8464214}\cite{Newton:2005:PPD:1038060.1038220} or it tries to prevent the estimation of these facial attributes while maintaining the recognition ability \cite{8272743} (soft-biometric privacy).
In the latter case, the solutions were based on image fusion, perturbations, and adversarial networks.

In order to suppress the gender of a face image, Suo et \textit{al.} \cite{5559505} proposed an approach that flips the estimated gender by decomposing the face image and replacing the facial components with similar parts of the opposite gender.
Othman and Ross presented an approach \cite{SuppressingGender} with the same goal.
They proposed a face morphing methodology that iteratively morphs two images and therefore, suppresses gender information at different levels.
However, this resulted in morphed images with many artefacts.

In \cite{DBLP:journals/corr/abs-1801-02480} and \cite{DBLP:journals/corr/RozsaGRB16},
adversarial images were created by using a fast flipping attribute technique, showed that it was able to fool their network in predicting binary facial attributes.
An incremental flipping approach was proposed by Mirjalili et \textit{al.} \cite{8272743} with the use of perturbations.
In \cite{kFacialAttributes}, imperceptible noise was used in order to suppress $k$ attributes at the same time.
However, this noise is trained to suppress attribute from only one specific neural network classifier and consequently, does not generalize.

In \cite{SemiAdNetworks}\cite{Mirjalili2018GenderPA}\cite{DBLP:journals/corr/abs-1905-01388}, Mirjalili et \textit{al.} proposed semi-adversarial networks consisting of a convolutional autoencoder, a gender classifier, and a face matcher.
It enhances the soft-biometric privacy on image level.
The autoencoder perturbs the input face image such that it minimizes gender classifier performance while trying to preserve the performance of the face matcher.
Training this supervised approach requires a large amount of data with the corresponding privacy-sensitive information.
However, the aim of privacy-enhancing technologies is to prevent a collection of privacy-sensitive attributes.
Moreover, this approach is limited at suppressing pre-defined attributes and thus, it is vulnerable to unseen function creep attacks.

Since most biometric representations are stored in templates rather than images \cite{DBLP:conf/ncc/DeyBBDCPS14}\cite{Stokkenes:2016:BAP:2947626.2951962}, privacy-enhancing approaches on template-level are presented in \cite{DBLP:conf/icb/Terh19}, \cite{Terhoerst2019AI} and \cite{DBLP:journals/corr/abs-1902-00334}.
These works further investigate the privacy performance in a more critical and challenging context of a function creep attacker that knows the systems privacy-mechanism.
While Terh\"{o}rst et \textit{al.} \cite{DBLP:conf/icb/Terh19} proposed an incremental variable elimination approach to eliminate privacy-risk features, Morales et \textit{al.} \cite{DBLP:journals/corr/abs-1902-00334} suppress attribute information via a modified triplet loss.
Both approaches aim at the suppression of very specific attributes which increases the risk of unseen attacks.

In \cite{Terhoerst2019AI}, this problem was tackled by proposing an unsupervised privacy-enhancing technique using similarity-sensitive noise transformations on template-level.
This unsupervised privacy-enhancing approach offers a more general privacy-protection not limited to a pre-defined attribute.
However, it struggles with achieving high suppression-rates while maintaining a high recognition ability.

In this work, we present an unsupervised privacy-enhancing face recognition approach working on template-level.
Using positive and negative template domains, it is able to achieve high privacy-sensitive attribute suppression rates in a more critical and challenging scenario of a function creep attacker that adapts his attacks to the systems privacy mechanism.
The achieved privacy-protection is, by design, not limited to the suppression of single attributes and further maintains a high recognition ability.

\section{Methodology}







Enhancing the soft-biometric privacy aims at preventing function creep attackers from reliably estimating privacy-risk characteristics.
This problem is further challenged by simultaneously maintaining a high recognition ability.
To solve these issues, we propose negative face recognition.
While in usual face recognition systems, the used templates describes the properties of an individual, our negative templates only contain complementary information and thus, describes properties that a person do not have.
We store only negative (reference) templates in a database and comparing it with positive (probe) templates by calculating their dissimilarity.
Due to the complementary nature of the compared templates, a high dissimilarity indicates that the templates belong to the same subjects and vice versa. 
Since the stored negative templates only contain some random complementary information, it prevents function creep attackers from successfully deducing privacy-sensitive information.
Further, it allows a more generalized soft-biometric privacy-protection that is, unlike previous works, not limited to the suppression of a pre-defined characteristic.
It is further a promissing candidate for template protection, as shown in a similar approach \cite{DBLP:journals/tdsc/ZhaoLLY18} for iris. It provides noninvertability, revocability, and nonlinkability, which are the key properties of template protection.
However, the template protection applicability is out of the scope of this work.

Since the idea of this work is to store only random complementary information of an individual in the database (in form of negative templates), the next section describes the enrolment process.
This is followed by a section of the adapted verification process, because the template comparison within our approach is dealing with complementary template versions.


\subsection{Enrolment phase}
In the enrolment phase, given a face image $I$, the corresponding face embedding $x$ is extracted from $I$.
Then, this embedding is transformed in the negative domain resulting in a negative template $t_-$ , which is stored in the database.
The generation of a negative template $t_-$ from a face embedding $x$ is done in three steps:
First, the face embedding $x$ is enlarged to get a higher-dimensional version $v$.
Second, $v$ is discretized to create a positive template $t_+$, and third, a negative template $t_-$ is generated from its positive complement by replacing each feature entry with a random value that does not match the original entry.

\subsubsection{Embedding enlargement}
\label{sec:EmbeddingComputation}

In the first step, the given face embedding $x$ is transformed to a higher-dimensional space while maintaining its recognition ability.
Therefore, a face recognition model, called enlargement-network, is trained to take the used face embedding $x$ as an input and outputs the higher-dimensional face embedding $v$ of size $L$.
The network and its training is described in Section \ref{sec:Enlargment}.
The enlargement step is necessary due to the fact that (a) the genuine/imposter decision is  based on the dissimilarity between a positive and a negative template and (b) the negative template generation is based on a randomized process.
If a positive and a negative template belong to different subjects, but are of low dimensionality, there is a higher chance that the negative template is very dissimilar from the positive one.
For increased dimensionalities, the positive and the negative templates share more similar feature entries and thus, increases the similarity.
In terms of positive-negative template comparison, a high similarity indicates an imposter comparison.
Consequently, high dimensional templates are needed for negative face recognition to reduce the recognition errors from the randomization process.

%
%
%

\subsubsection{Embedding discretization}
\label{sec:EmbeddingDiscetization}
In the second step, the enlarged embedding $v$ is feature-wise discretized into $k$ bins.
The $k$ bins were chosen beforehand on the enlarged training data using a quantile strategy that divides each feature range into $k$ bins such that every bin contains an approximately equal number of samples.
Following this binning ranges, each feature entry of $v$ is replaced with the value $l\in \mathcal{K}=\left\lbrace 1 \dots, k\right\rbrace$ of its corresponding bin.
This results in a discretized positive template $t_+ \in \mathcal{K}^L$.
Discrete features are required for the feature-wise computation of complementary feature sets that is needed in next step of the negative template generation.


\subsubsection{Negative template generation}
\label{sec:NegativeTemplateGeneration}
The third step replaces each feature entry of the positive template $t_+$ with a random value from the complementary feature set.
This results in a negative template which contains facial properties that the person does not possess and thus, it is hard to estimate the soft-biometrics of that individual.
Given a positive template $t_+ \in \mathcal{K}^L$, a negative associated template $t_-$ is generated feature-wise.
This is done by replacing each feature entry of $t_+$ with a randomly chosen value from $\mathcal{K}$ that does not match the original entry.
To be precise, for each component $i$ the negative representation $t_-^{(i)} \in \mathcal{K} \setminus \lbrace t_+^{(i)} \rbrace$ is given by a randomly chosen value from the complement set $\mathcal{K} \setminus \left\lbrace t_+^{(i)} \right\rbrace$.
This results in the negative template $t_-$, which is stored in the gallery database.


\subsection{Verification phase}
In the verification phase, a negative (reference) template stored in the database is compared with a positive (probe) template from a captured individual.
In order to verify a persons identity with our negative face recognition approach, (1) the positive probe template and the negative reference template have to be allocated and (2) the templates are compared against each other to determine a comparison score.
This comparison score is used to make a verification decision.

\subsubsection{Template allocation}
Given an input face image from an individual, first, its embeddings (Section \ref{sec:EmbeddingComputation}) have to be computed and second, discretized (Section \ref{sec:EmbeddingDiscetization}) to obtain the positive (probe) template $t_+$.
The negative (reference) template $t_-$ is loaded from the database.
The positive and the negative template can then be compared.

\subsubsection{Positive-negative template comparison}

In order to compute a comparison score between the positive and the negative template $t_+$ and $t_-$, we utilize a normalized hamming-like distance
\begin{align}
\textit{NHD}(t_+, t_-) = 1-\dfrac{1}{|t_+ |} \sum_{i=1}^{|t_+|} \delta \left( t_+^{(i)}, t_-^{(i)} \right) .
\end{align} 
The delta function $\delta(a,b)$ returns 1 if $a$ equals $b$ and 0 otherwise.
The size of the templates is defined by $|t_+|=L$.
The \textit{NHD} measures the dissimilarity of $t_+$ and $t_-$ and, due to the complement nature of positive and negative templates, it can be directly utilized as a comparison score.
Since the positive template defines properties of the corresponding individual, while the negative template describes properties that the individual does not contain, a larger (\textit{NHD}) distance represents a higher probability that the templates originate from the same subject and vice versa.

\subsection{Discussion about the gain of privacy}
Our negative face recognition approach makes a face recognition system less vulnerable to function creep attacks in cases where attackers get access to the stored data.
Since only negative templates are stored, the information about an individual is limited to the deeply-encoded description of complementary nature.
This enables our solution to offer a more comprehensive privacy-protection that is not limited to single pre-defined attributes.

However, in the case of function creep attackers getting access to multiple negative templates that were created from the same positive templates, a statistical analysis might enable a reconstruction of the positive template.
Consequently, in this special case, a reconstruction and thus, a reliable privacy-sensitive attribute estimation might be possible.
To prevent this attack strategy, we recommend the use of differently-trained enlargement-networks for different databases.
This prevents that the generation of negative templates from the same positive template and thus, circumvent this statistical analysis-based attack strategy even in the case of attackers getting access to multiple negative face databases.

\subsection{Theoretical foundation}
\label{sec:Theory}
Since our approach is based on a randomized process in the template generation, we provide a statistical reasoning for the negative-positive comparison performance.
Given a theoretical or empirical score distribution of genuine and imposter scores, this allows to predict the negative face recognition performance including the probabilities of falsely rejected and falsely accepted subjects.
Consequently, optimal hyperparameters can be chosen, as well as large-scale experiments can be simulated, without the computational costs for sophisticated experiments.

Given two positive templates $t_+^A,t_+^B \in \mathcal{K}^L$ with a distance of,
\begin{align}
\textit{HD}(t_+^A,t_+^B)=\sum_{i=1}^{|t_+|} \delta \left( t_+^{(i)}, t_-^{(i)} \right)=D,
\end{align}
then, the probability of this distance in the negative domain,
\begin{align}
\textit{HD}(t_-^A,t_+^B)=D' \quad \text{with} \quad D'\in\left[L-D, L\right]
\end{align}
follows a Bernoulli distribution and is given by,
\begin{align}
Pr\left[D'|D \right] = {D \choose \mu(D')} \left(\dfrac{k-2}{k-1}\right)^{\mu(D')} \left(\dfrac{1}{k-1}\right)^{D-\mu(D')}. \label{eq:Proba_PosNeg}
\end{align}
The Bernoulli distribution can be assumed, since only entries of equal values contribute to the distance and the state of this entries is given by a fixed probability.
The number of bins in this equation is described by $k = |\mathcal{K}|$ and $\mu(D')$ is given by
\begin{align}
\mu(D') = D'-\left(L-D \right),
\end{align}
the number of entries that have to be flipped to the same entry in order to achieve the determined distance of $D'-D$. 
Based on our negative template generation principle (Section \ref{sec:NegativeTemplateGeneration}), colliding bin labels $t_+^A$ and $t_+^B$ in the positive domain, will not collide in the negative domain and thus, will contribute to the distance.
In order to achieve a distance of $\textit{HD}(t_-^A,t_+^B)=D'$ in the negative domain, $\mu(D')$ bin labels have to be flipped such that they will contribute to the distance calculation.
The probability for such a single flip is given by $\frac{k-2}{k-1}$ and thus, the collision probability is given by $\frac{1}{k-1}$. 

Given two positive templates $t_+^A,t_+^B$ with a distance of $\textit{HD}(t_+^A,t_+^B)=D$, then Equation \ref{eq:Proba_PosNeg} gives the probability for the two templates to have a hamming distance of $D'$ if one of the templates is in the negative domain.

\section{Experimental setup} 
\label{sec:ExperimentalSetup}

\subsection{Database}
In order to evaluate and compare our approach under both controlled and uncontrolled conditions, we conducted experiments on the public available ColorFeret \cite{ColorFERET} and Adience \cite{Eidinger:2014:AGE:2771306.2772049} databases.
ColorFeret \cite{ColorFERET} consists of 14,126 images from 1,199 different individuals with different poses under controlled conditions.
A variety of face poses, facial expressions, and lighting conditions are included in the dataset.
The Adience dataset \cite{Eidinger:2014:AGE:2771306.2772049} consists of 26,580 images from over 2,284 different subjects under uncontrolled imaging conditions.
Adience contains additional information about gender and age.
ColorFeret also provides labels regarding the subjects ethnicities.
We choose these databases because they were captured under controlled and uncontrolled conditions and provide information of soft-biometric attributes and information about the identities.
This allows to deeply investigate the privacy-enhancing technologies by analysing the recognition performance, as well as the suppression performance of privacy-sensitive attributes.


\subsection{Evaluation metrics}
Enhancing soft-biometric privacy describes a trade-off between the desired degradatioin of the attribute estimation performance by function creep attackers and the desired preservation the recognition ability.
In this work, we report our verification performances in terms of false non-match rate (FNMR) at fixed false match rates (FMR).
We also report the equal error rate (EER), which equals the FMR at the threshold where FMR = 1$-$FNMR.
This acts as a single-value indicator of the verification performance.
In order to evaluate the attribute suppression performance, we report our results in terms of attribute classification accuracy on balanced test labels and in terms of attribute suppression rates.
The suppression rate describes reduction of the attribute-prediction accuracy of the templates without privacy-enhancement in comparison to the original templates.

\subsection{Basic face recognition model}
The proposed negative face recognition approach builds on arbitrary face embeddings.
In this work, we utilize the widely used FaceNet model\footnote{\url{https://github.com/davidsandberg/facenet}} \cite{DBLP:journals/corr/SchroffKP15}, which was pretrained on MS-Celeb-1M \cite{DBLP:journals/corr/GuoZHHG16}.
In order to extract an embedding of a face image, the image is aligned, scaled, and cropped as described in \cite{Kazemi2014OneMF} and then passed into the model.
The output of this network is a 128-dimensional embedding.
The comparison of two such embeddings is performed using cosine-similarity.

\subsection{Enlargement-network training}
\label{sec:Enlargment}

\begin{figure}[h]
\centering
\includegraphics[width=0.45\textwidth]{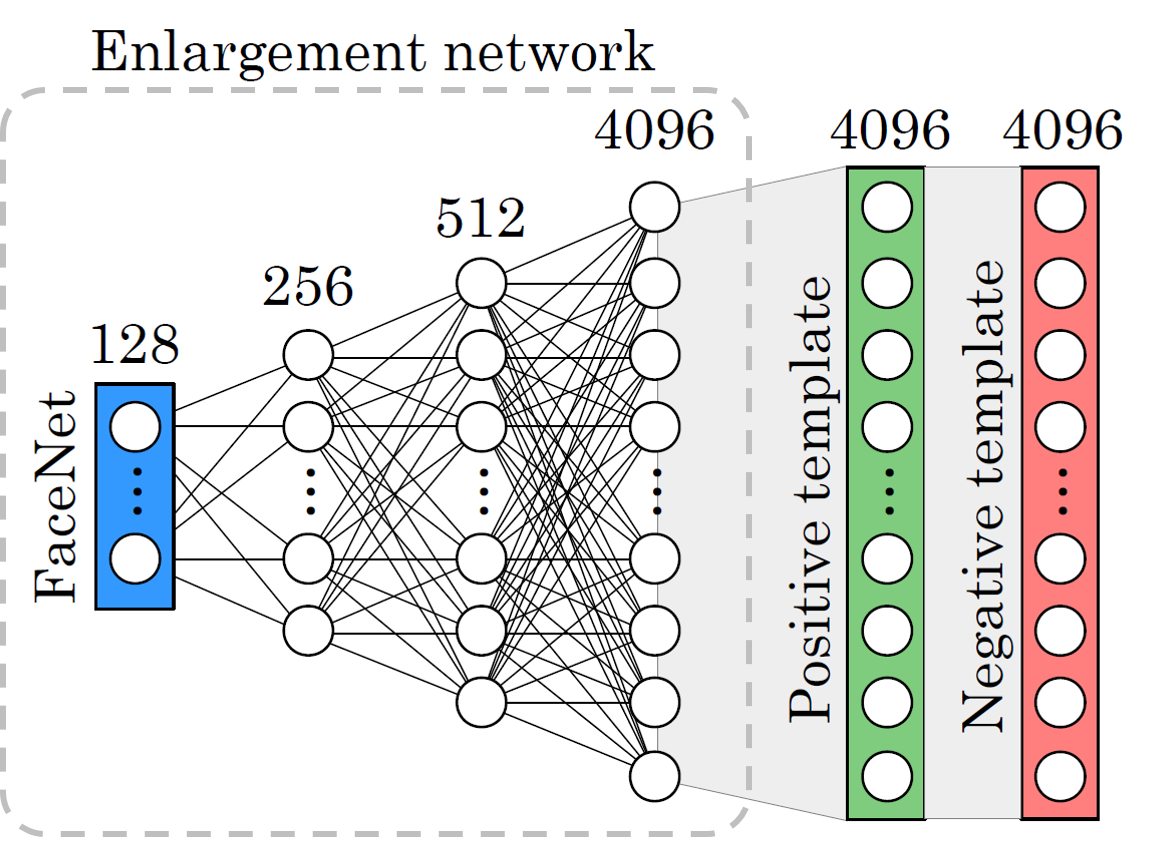}
\caption{Enlargement-network and positive/negative template generation: the enlargement-network structure is shown without softmax layer. Given a face embedding (FaceNet) a larger representation of this embedding is computed. A positive template is created by discretisation and replacing each feature entry with an item its complementary set, a negative template is generated.}
\label{fig:NN}
\end{figure}

Our approach requires high-dimensional face embeddings, in order to create discriminative negative templates.
As described in Section \ref{sec:EmbeddingComputation}, an enlargement-network is used to expand the low-dimensional face embeddings to size $L$.
This network has an input size of 128, corresponding to the used face embeddings, and an output size of $L=4096$.
It consists of three layers with 256, 512, and 4096 neurons, and is shown in Figure \ref{fig:NN}.
The first layers are activated by a ReLU function, while the forth layer holds a tanh activation, such that the output-features are within the range of $[-1,1]$.
To train the network, a softmax layer is added to classify the identities in the test set with a binary cross-entropy loss.
The training is done with an AdaDelta optimizer (learning rate $lr=0.5$) over 50 epochs of training.
Dropout ($p=0.5$) \cite{JMLR:v15:srivastava14a} and Batchnormalization \cite{Ioffe:2015:BNA:3045118.3045167} is applied on every layer.
After the training the softmax layer is removed.

\subsection{Function creep attacks}
For the evaluation of the attribute suppression, we simulate the critical scenario of a function creep attacker that adapts to the systems privacy mechanism. 
The adaptation step is done by training (function creep) classifiers on the transformed (normalized and scaled) templates to predict the privacy-sensitive attributes.
These classifiers include random forest (RF), support vector machines (SVM), k-nearest neighbors (kNN), and logistic regression (LogReg).
The hyperparameters of these classifiers are fine-tuned with Bayesian optimization.

\subsection{Baseline approaches}
In Section \ref{sec:RelatedWork}, we mentioned that many privacy-enhancing methods were proposed that manipulate the face images itself using supervised approaches.
However, most biometric systems store face templates instead of images \cite{DBLP:conf/ncc/DeyBBDCPS14}\cite{Stokkenes:2016:BAP:2947626.2951962} and furthermore, supervised approaches are vulnerable to attacks on attributes that were unconsidered during training.
Therefore we proposed an unsupervised privacy-enhancing approach working on template-level and compare it against two state-of-the-art solutions with the same working principles.
In this work, we use similarity-sensitive noise transformations \cite{Terhoerst2019AI} as baselines.
More precisely, we compare our proposed NFR approach against cosine-sensitive noise (CSN) and euclidean-sensitive noise (ESN).

%
%

We calibrate the hyperparameters of these baselines in such a way that they reach similar  verification EER performances.
By doing so it is possible to fairly compare these methods in terms of suppression rates.
For all experiment scenarios, subject-disjoint 5-fold cross-validation is utilized.
The performance over all folds is reported as the average performance and its standard deviation.

\subsection{Investigations}
The investigations of this work are divided in five parts:
\begin{enumerate}[1.]
\item We show the need for a privacy-enhancing technology by demonstrating that there is a significant leakage of privacy-sensitive information from face templates on both databases.
\item We analyse the face verification performance of our privacy-enhancing solution to check to which degree the recognition ability is maintained and compare it with previous works.
\item We investigate the attribute suppression performance of our solution and the baselines in the critical scenario of a function creep attacker that adapts to the systems privacy mechanism.
This evaluates the soft-biometric privacy protection.
\item We analyse the parameter space of our solution to provide a deeper understanding of our solution.
\item Lastly, we provide a empirical validation of the theoretical reasoning and validate its correctness.
\end{enumerate}

%


\section{Results}

\subsection{Analysis of the function creep performance}

\begin{table*}[h]
\renewcommand{\arraystretch}{1.2}
\centering
\caption{Attribute prediction performance on original and positive templates (without privacy-enhancement). The prediction accuracies of four function creep estimators are shown on two databases.
A function creep attacker would be able to predict the soft-biometric attributes with high accuracies demonstrating the need for privacy-enhancement.}
\label{tab:attributePrediction}
\begin{tabular}{lllcccc}
\Xhline{2\arrayrulewidth} 
Representation & Dataset & Attribute & RF & SVM & kNN & LogReg          \\
\hline
Original & ColorFeret & Gender   & 93.37\%      $\pm$ 1.22\%      & 96.61\%       $\pm$ 1.02\%      & 97.30\%       $\pm$ 0.39\%         & 95.74\%          $\pm$ 1.14\% \\
 & & Age      & 49.17\%      $\pm$ 2.40\%      & 57.40\%       $\pm$ 2.63\%      & 47.71\%       $\pm$ 2.06\%         & 57.12\%          $\pm$ 2.91\% \\
 & & Race     & 82.21\%      $\pm$ 1.07\%      & 88.73\%       $\pm$ 1.17\%      & 85.10\%       $\pm$ 2.58\%        & 88.03\%          $\pm$ 1.46\% \\
\cmidrule{3-7}
 & Adience & Gender & 82.78\% $\pm$ 1.79\% & 84.16\% $\pm$ 2.34\% & 84.91\% $\pm$ 2.45   & 82.43\% $\pm$ 2.78\% \\
 & & Age    & 53.27\% $\pm$ 4.08\% & 60.36\% $\pm$ 4.10\% & 51.31\% $\pm$ 3.08\%   & 58.26\% $\pm$ 4.93\% \\
\hline
Positive (b=3) & Colorferet & Gender & 90.05\% $\pm$ 2.08\% & 96.92\% $\pm$ 0.86\% & 94.89\% $\pm$ 0.69\% & 93.45\% $\pm$ 0.86\% \\
 &  & Age & 44.47\% $\pm$ 2.78\% & 53.51\% $\pm$ 1.69\% & 45.99\% $\pm$ 1.69\% & 46.95\% $\pm$ 1.19\% \\
 &  & Race & 79.73\% $\pm$ 0.84\% & 87.52\% $\pm$ 1.59\% & 84.09\% $\pm$ 2.33\% & 84.46\% $\pm$ 1.92\% \\
 \cmidrule{3-7}
 & Adience & Gender & 77.73\% $\pm$ 1.82\% & 87.88\% $\pm$ 2.72\% & 82.89\% $\pm$ 2.64\% & 80.48\% $\pm$ 2.62\% \\
 &  & Age & 48.48\% $\pm$ 2.15\% & 59.51\% $\pm$ 3.60\% & 48.90\% $\pm$ 3.03\% & 49.06\% $\pm$ 3.25\% \\
 \hline
Positive (b=4) & Colorferet & Gender & 90.38\% $\pm$ 2.51\% & 97.12\% $\pm$ 0.59\% & 95.57\% $\pm$ 0.62\% & 93.79\% $\pm$ 0.72\% \\
 &  & Age & 45.46\% $\pm$ 1.33\% & 54.54\% $\pm$ 1.23\% & 47.29\% $\pm$ 1.07\% & 48.20\% $\pm$ 1.06\% \\
 &  & Race & 80.46\% $\pm$ 0.88\% & 87.66\% $\pm$ 1.82\% & 83.93\% $\pm$ 2.05\% & 85.10\% $\pm$ 1.53\% \\
 \cmidrule{3-7}
 & Adience & Gender & 76.71\% $\pm$ 2.01\% & 87.38\% $\pm$ 1.89\% & 82.95\% $\pm$ 1.68\% & 79.43\% $\pm$ 1.75\% \\
 &  & Age & 50.43\% $\pm$ 2.62\% & 60.74\% $\pm$ 2.86\% & 51.10\% $\pm$ 2.66\% & 51.70\% $\pm$ 2.00\%\\
 \Xhline{2\arrayrulewidth} 
\end{tabular}
\end{table*}

Table \ref{tab:attributePrediction} shows the attribute prediction performance of three privacy-sensitive attributes in a scenario without privacy-preservation.
The performance of four function creep classifiers is shown under controlled (ColorFeret) and uncontrolled (Adience) circumstances using the original and positive embeddings.
Especially gender and race can be determined with very high accuracies.
This holds true for the original embeddings as well as the (high dimensional and discrete) positive embeddings.
The table demonstrates that there is a significant information leakage of privacy-sensitive information from face templates and thus, a great need for privacy-enhancing technologies.

\subsection{Face verification performance}

\begin{table*}[h]
\renewcommand{\arraystretch}{1.2}
\centering
\caption{Face recognition performance on ColorFeret. The original face recognition performance is compared against three privacy-enhancing approaches, our proposed negative face recognition approach, cosine-sensitive noise (CSN), and euclidean-sensitive noise (ESN).}
\label{tab:ColorFeret-Recognition}
\begin{tabular}{lrrrr}
\Xhline{2\arrayrulewidth} 
                 & \multicolumn{1}{c}{FNMR@$10^{-2}$\,FMR}   & \multicolumn{1}{c}{FNMR@$10^{-3}$\,FMR}   &              \multicolumn{1}{c}{EER} & \\
                 \hline
Original         & 3.65\% $\pm$ 0.95\% & 14.22\% $\pm$ 3.49\% & 1.97\% $\pm$ 0.21\%                    &   \\ \\
Ours ($k=3$)  & 6.50\%  $\pm$ 1.10\% & 18.32\% $\pm$ 4.23\% & 3.18\% $\pm$ 0.20\%                     \\
CSN ($\Theta=0.80$)   & 7.61\%  $\pm$ 1.01\% & 23.54\% $\pm$ 4.42\% & 3.25\% $\pm$ 0.19\%                     \\
ESN ($r=0.75$)  & 7.47\%  $\pm$ 1.12\% & 23.55\% $\pm$ 3.93\% & 3.21\% $\pm$ 0.25\%                     \\ \\
Ours ($k=4$)  & 8.65\%  $\pm$ 1.22\% & 20.26\% $\pm$ 2.81\% & 4.15\% $\pm$ 0.40\%                     \\
CSN ($\Theta=0.73)$  & 11.18\% $\pm$ 1.24\% & 32.16\% $\pm$ 5.03\% & 4.20\% $\pm$ 0.23\%                     \\
ESN ($r=0.93$)  & 11.56\% $\pm$ 0.99\% & 33.01\% $\pm$ 4.54\% & 4.24\% $\pm$ 0.19\%                     \\
\Xhline{2\arrayrulewidth} 
\end{tabular}
\end{table*}

\begin{table*}[h]
\renewcommand{\arraystretch}{1.2}
\centering
\caption{Face recognition performance on Adience. The original face recognition performance is compared against three privacy-enhancing approaches, our proposed negative face recognition approach, cosine-sensitive noise (CSN), and euclidean-sensitive noise (ESN).}
\label{tab:Adience-Recognition}
\begin{tabular}{lrrrr}
\Xhline{2\arrayrulewidth} 
                 & \multicolumn{1}{c}{FNMR@$10^{-2}$\,FMR}   & \multicolumn{1}{c}{FNMR@$10^{-3}$\,FMR}   &              \multicolumn{1}{c}{EER} & \\
                 \hline
Original         & 13.68\%              $\pm$ 5.24\%              & 45.71\%                $\pm$ 6.88\%  & 3.83\%   $\pm$ 0.72\% \\
 \\
Ours ($k=3$)  & 13.42\%               $\pm$ 4.79\%              & 43.14\%                $\pm$ 9.14\% & 4.43\%   $\pm$ 0.80\% \\
CSN ($\Theta=0.84$)  & 19.36\%               $\pm$ 6.30\%              & 59.04\%                $\pm$ 6.90\%  & 4.48\%   $\pm$ 0.75\% \\
ESN ($r=0.62$)  & 18.60\%               $\pm$ 5.76\%              & 57.56\%                $\pm$ 6.59\%  & 4.49\%   $\pm$ 0.72\% \\
 \\
Ours ($k=4$)  & 16.35\%               $\pm$ 4.20\%              & 47.93\%                $\pm$ 8.69\%  & 5.44\%   $\pm$ 0.78\% \\
CSN ($\Theta=0.74$) & 28.29\%               $\pm$ 7.49\%              & 71.79\%                $\pm$ 6.59\%  & 5.57\%   $\pm$ 0.80\% \\
ESN ($r=0.88$)  & 28.16\%               $\pm$ 6.91\%              & 71.27\%                $\pm$ 5.73\%  & 5.49\%   $\pm$ 0.70\%               \\
\Xhline{2\arrayrulewidth} 
\end{tabular}
\end{table*}

In Table \ref{tab:ColorFeret-Recognition} and \ref{tab:Adience-Recognition}, the recognition performance of the baseline approach is shown in comparison to state-of-the-art \cite{Terhoerst2019AI} and our approach.
In order to make a fair comparison of the attribute suppression analysis, the hyperparameters of both unsupervised state-of-the-art approaches, CSN and ESN, are calibrated such that it matches the EER of our approach for $k=3$ and $k=4$ bins.
In Table \ref{tab:ColorFeret-Recognition}, the recognition performance is shown for the ColorFeret database.
While the EER of templates without privacy-enhancement is about 2\%, our approach with $k=3$ ($k=4$) bins leads to an EER around 3\% (4\%). 
Even if CSN and ESN are calibrated to have a comparable EER, their FNMR for low FMR is significantly higher than our approach.
In Table \ref{tab:Adience-Recognition}, the recognition performance is shown for the Adience database.
It is observed that the recognition performance for our approaches ($k=3$ and $k=4$) is very close to the original performance, while the CSN and ESN show a strongly degraded performance.
While CSN and ESN are based on noise injections that leads to a partial identity loss, our approach is based on a complementary representations, which keeps the identity information,
but transforms it in an irreversible manner.

\begin{figure*}[h]
\centering
  \subfloat[ColorFeret \label{fig:ROC_ColorFeret}]{%
       \includegraphics[width=0.49\textwidth]{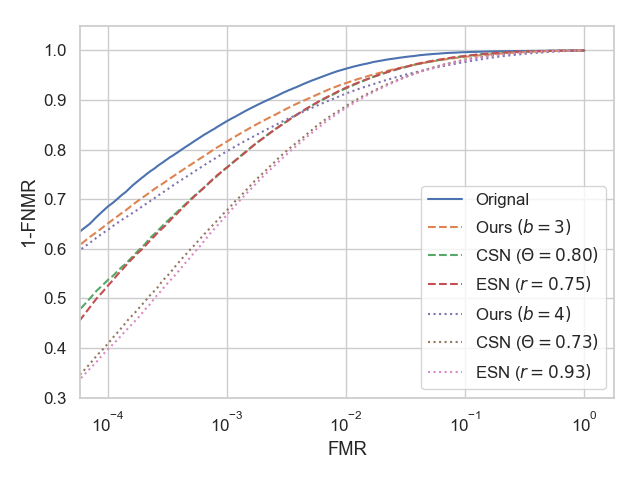}} \hspace{1mm}
  \subfloat[Adience \label{fig:ROC_Adience}]{%
       \includegraphics[width=0.49\textwidth]{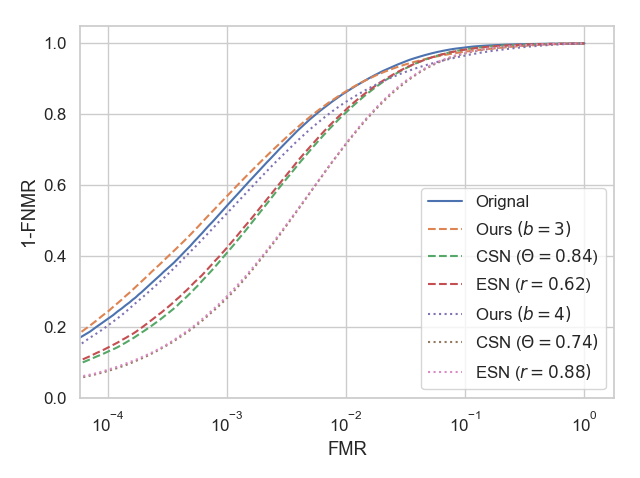}}
\caption{Face recognition performance comparing the performance of the original templates, our approach and related work. Our solution is able to maintain the verification performance to a higher degree then previous works.}
\label{fig:ROC}
\end{figure*}

To get a more detailed look in the recognition performances over a wider range of decision thresholds, Figure \ref{fig:ROC} shows ROC curves on both datasets.
In Figure \ref{fig:ROC_ColorFeret}, the performance is shown under controlled face image capture conditions, while in Figure \ref{fig:ROC_Adience} the same is shown under uncontrolled conditions.
In both cases, it can be observed that recognition performance is very close to the performance of the original representations, while the CSN and ESN shows a strongly degraded performance.
Especially under uncontrolled conditions (Figure \ref{fig:ROC_Adience}) the performance even surpasses the performance of the original representations by a small amount due to its error correction ability.
This demonstrates, in contrast to previous work, that our solution is able to maintains identity information to a large degree.

\subsection{Privacy-sensitive attribute suppression}

\begin{table*}[h]
\renewcommand{\arraystretch}{1.2}
\setlength{\tabcolsep}{5pt}
\centering
\caption{Attribute suppression performance on the ColorFeret and Adience databases. The gender, age, and race suppression rates are shown for four function creep estimators.
The highest suppression rates are highlighted.}
\label{tab:AttributeSuppression}
\begin{tabular}{llrrrrrrrrrrrr}
\Xhline{2\arrayrulewidth} 
 & &   \multicolumn{4}{c}{Gender suppression rate}      &     \multicolumn{4}{c}{Age suppression rate}   & \multicolumn{4}{c}{Race suppression rate}   \\
 \cmidrule(lr){3-6} \cmidrule(lr){7-10} \cmidrule(lr){11-14}
              &          & \multicolumn{1}{c}{RF}      & \multicolumn{1}{c}{SVM}     & \multicolumn{1}{c}{kNN}     & \multicolumn{1}{c}{LogReg}               & \multicolumn{1}{c}{RF}      & \multicolumn{1}{c}{SVM}     & \multicolumn{1}{c}{kNN}    & \multicolumn{1}{c}{LogReg}                & \multicolumn{1}{c}{RF}      & \multicolumn{1}{c}{SVM}    & \multicolumn{1}{c}{kNN}    & \multicolumn{1}{c}{LogReg}  \\
                        \hline
\parbox[t]{1mm}{\multirow{6}{*}{\rotatebox[origin=c]{90}{ColorFeret}}} & Ours ($k=3$)         & \textbf{22.2\%} & \textbf{4.5\%}  & \textbf{5.5\%}  & \textbf{6.9\%}                & \textbf{30.2\%} & \textbf{11.1\%} & \textbf{9.3\%} & \textbf{26.1\%}                & \textbf{14.6\%} & \textbf{4.1\%} & \textbf{4.1\%} & \textbf{7.8\%}  \\
&CSN ($\Theta=0.80$)          & 7.3\%  & 3.6\%  & 1.5\%  & 4.1\%               & 10.6\% & 6.0\% & 2.4\% & 6.7\%                 & 6.1\%  & 2.3\% & 0.8\% & 2.5\% \\
&ESN ($r=0.75$)        & 7.8\%  & 3.8\%  & 1.6\%  & 3.9\%                & 9.7\%  & 5.9\%  & 1.2\% & 7.4\%                 & 5.9\%  & 2.3\% & 0.4\% & 2.8\%  \vspace{2mm} \\
&Ours ($k=4$)         & \textbf{28.4\%} & \textbf{7.4\% } & \textbf{14.4\%} & \textbf{10.9\%}               & \textbf{34.2\%} & \textbf{13.7\%} & \textbf{9.4\%} & \textbf{31.2\%}                & \textbf{21.8\%} & \textbf{7.6\%} & \textbf{9.3\%} & \textbf{12.0\%} \\
&CSN ($\Theta=0.73$)       & 11.0\% & 5.0\% & 2.6\%  & 5.3\%               & 12.1\% & 8.6\% & 4.2\% & 9.6\%                & 8.5\% & 3.7\% & 1.6\% & 4.3\% \\
&ESN ($r=0.93$)       & 10.3\% & 5.4\%  & 3.5\%  & 5.8\%                & 13.5\% & 8.1\%  & 4.8\% & 8.8\%                 & 8.7\%  & 3.6\% & 2.6\% & 4.0\%  \\
                        \hline 
\parbox[t]{1mm}{\multirow{6}{*}{\rotatebox[origin=c]{90}{Adience}}} & Ours ($k=3$)         & \textbf{26.1\%} & \textbf{4.3\%}  & \textbf{6.8\%}  & \textbf{12.3\%}              & \textbf{28.7\%} & \textbf{10.3\%} & \textbf{8.0\%}  & \textbf{25.4\%} & - & - & - & -\\
&CSN ($\Theta=0.84$)        & 8.8\% & 2.8\%  & 2.5\%  & 4.9\%                & 10.3\% & 4.9\% & 5.7\% & 5.8\% & - & - & - & - \\
&ESN ($r=0.62$)      & 7.0\%  & 4.3\%  & 2.5\%  & 5.2\%                & 10.1\% & 4.6\%  & 6.5\%  & 5.0\%  \vspace{2mm} & - & - & - & - \\
&Ours ($k=4$)         & \textbf{32.6\%} & \textbf{11.0\%} & \textbf{19.9\%} & \textbf{18.0\%} & \textbf{36.3\%} & \textbf{14.1\%} & \textbf{16.9\%} & \textbf{29.8\%}  & - & - & - & -\\
&CSN ($\Theta=0.74$)     & 14.6\% & 6.2\% & 6.2\% & 7.6\%                & 16.6\% & 9.1\% & 12.2\% & 9.8\%  & - & - & - & -\\
&ESN ($r=0.88$)        & 14.1\% & 7.2\%  & 6.5\%  & 7.7\%                & 15.1\% & 8.3\%  & 11.9\% & 8.4\%   & - & - & - & -\\
\Xhline{2\arrayrulewidth} 
\end{tabular}
\end{table*}

In order to compare the soft-biometrics privacy-enhancement, Table \ref{tab:AttributeSuppression} shows the suppression rates for four classifiers on three privacy-sensitive attributes.
The attribute suppression performances of our approach is shown and compared with state-of-the-art approaches (CSN, ESB) \cite{Terhoerst2019AI} calibrated to the same verification EER.
In \cite{Terhoerst2019AI}, CSN showed significantly better performance than ESN, especially in suppressing attribute prediction performance for SVM and LogReg.
However, CSN transforms each feature vector to a random length $r \in \left[1,100\right]$, which makes it hard to handle for classifiers such as SVM and LogReg.
This is not the case in our experiments, since we simulated a committed function creep attacker that does not only train on transformed data, but also rescales the feature vectors to unit-length.
This prevents classifiers, such as SVM and LogReg, from unstable estimations. 
On both databases, our solution achieves relatively high suppression rates on all classifiers and all attributes.
Generally, our privacy-enhancement approach leads to 2-4 times higher suppression rates compared to previous work under different attack mechanisms and attributes.

\subsection{Investigation of the parameter space}

\begin{figure*}[h!]
\centering
  \subfloat[Gender]{%
       \includegraphics[width=0.33\textwidth]{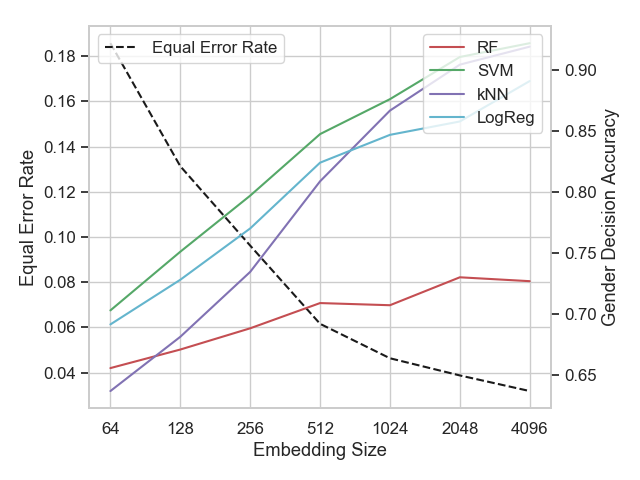}}
    \hfill
  \subfloat[Age]{%
       \includegraphics[width=0.33\textwidth]{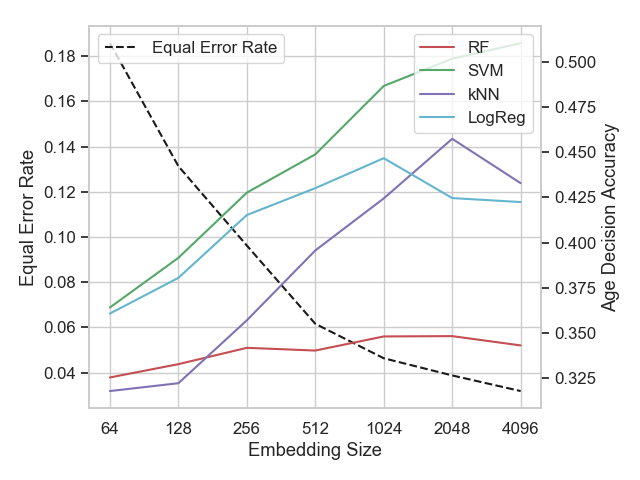}}
    \hfill
  \subfloat[Race]{%
       \includegraphics[width=0.33\textwidth]{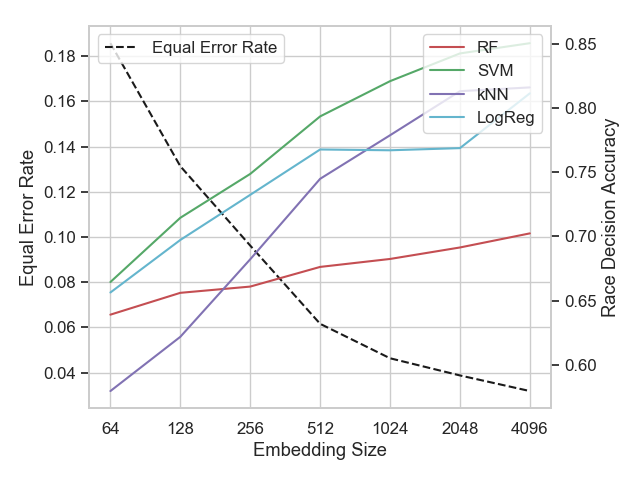}}
\caption{On ColorFeret the face verification EER and the attribute estimation performances of function creep estimators are shown for different embedding sizes and a fixed bin size of $k=3$. The estimation performance is analysed for the attributes gender, age, and race.}
\label{fig:EER_Attr_estimation_b3_ColorFeret}
\end{figure*}

\begin{figure*}[h!]
\centering
  \subfloat[Gender]{%
       \includegraphics[width=0.33\textwidth]{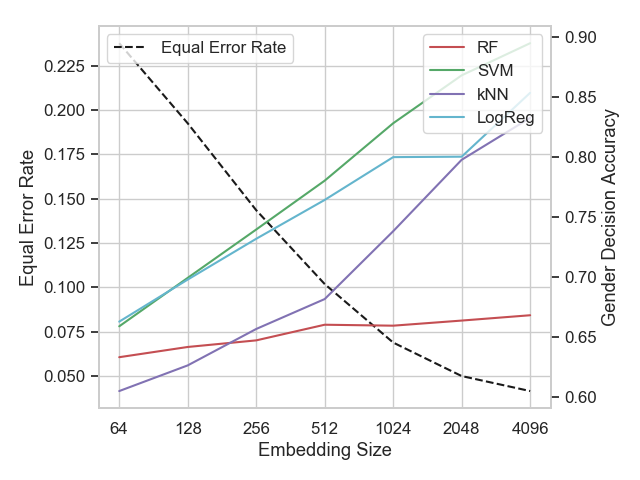}}
    \hfill
  \subfloat[Age]{%
       \includegraphics[width=0.33\textwidth]{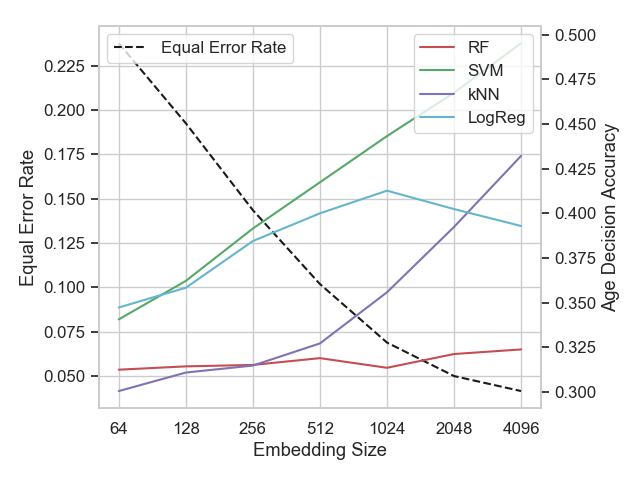}}
    \hfill
  \subfloat[Race]{%
       \includegraphics[width=0.33\textwidth]{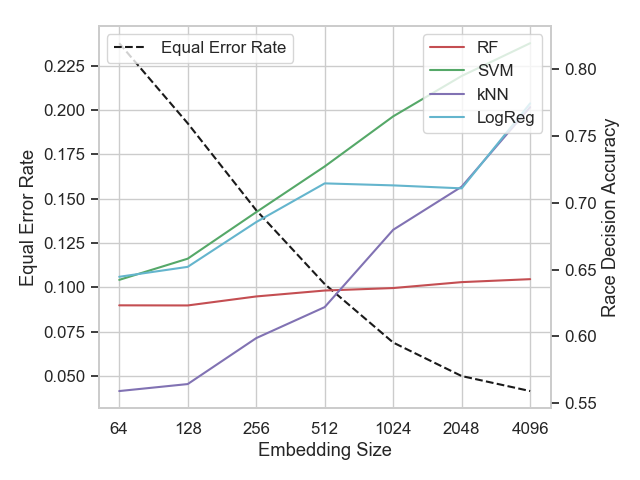}}
\caption{On ColorFeret the face verification EER and the attribute estimation performances of function creep estimators are shown for different embedding sizes and a fixed bin size of $k=4$. The estimation performance is analysed for the attributes gender, age, and race.}
\label{fig:EER_Attr_estimation_b4_ColorFeret}
\end{figure*}

\begin{figure*}[h]
\centering
  \subfloat[Gender]{%
       \includegraphics[width=0.49\textwidth]{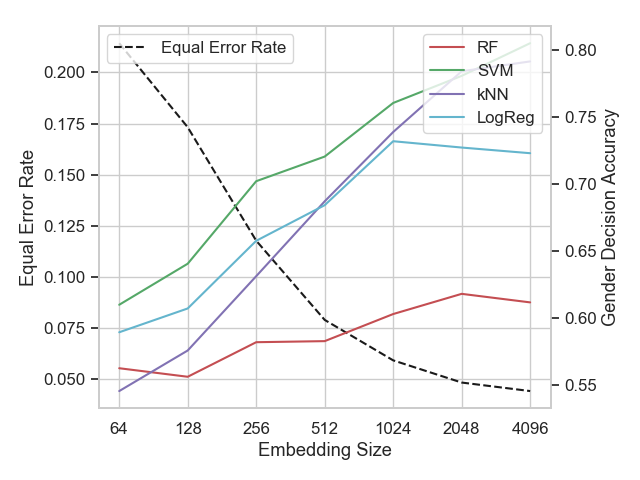}}
       \hfill
  \subfloat[Age]{%
       \includegraphics[width=0.49\textwidth]{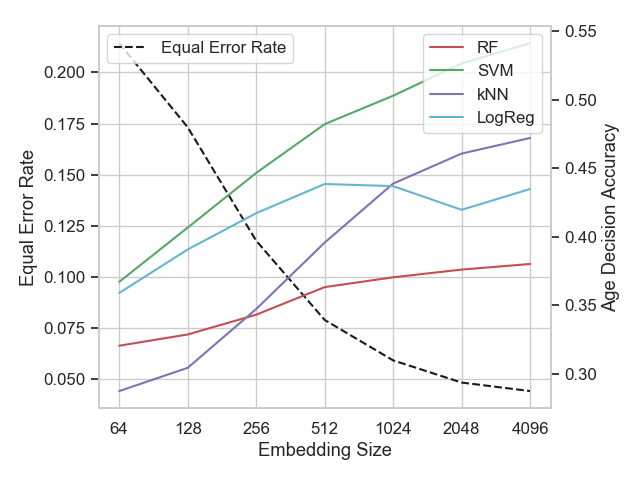}}
\caption{On Adience the face verification EER and the attribute estimation performances of function creep estimators are shown for different embedding sizes and a fixed bin size of $k=3$. The estimation performance is analysed for the attributes gender and age.}
\label{fig:EER_Attr_estimation_b3_Adience}
\end{figure*}

\begin{figure*}[h]
\centering
  \subfloat[Gender]{%
       \includegraphics[width=0.49\textwidth]{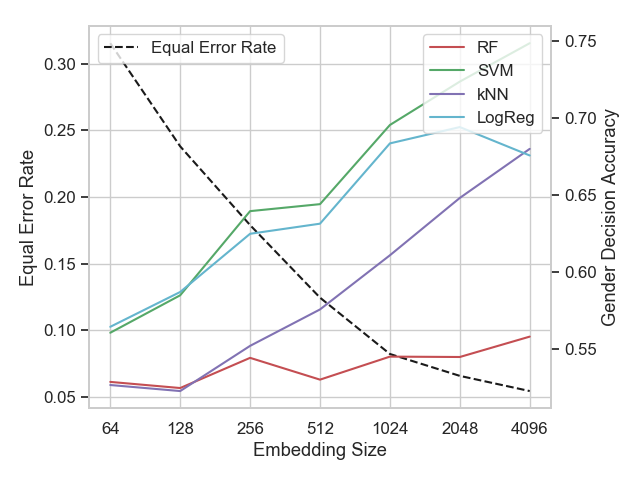}}
       \hfill
  \subfloat[Age]{%
       \includegraphics[width=0.49\textwidth]{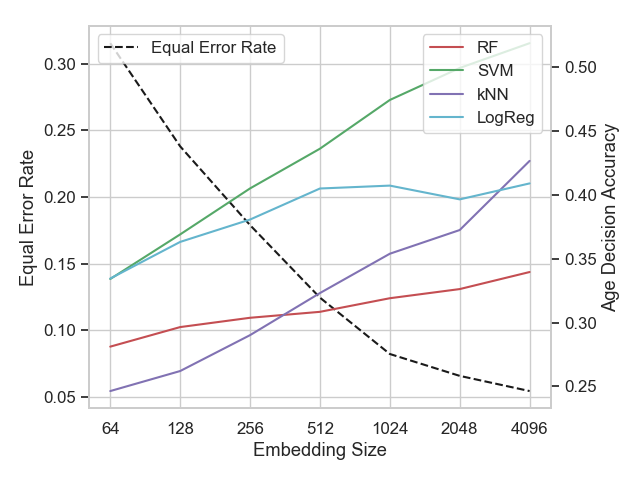}}
\caption{On Adience the face verification EER and the attribute estimation performances of function creep estimators are shown for different embedding sizes and a fixed bin size of $k=4$. The estimation performance is analysed for the attributes gender and age.}
\label{fig:EER_Attr_estimation_b4_Adience}
\end{figure*}

In the following, the parameter space is analysed to increase the understanding of our solutions behaviour.
More precisely, the two parameters of our solution, the template size $L$ and the number of bins $k$, are varied and for every parameter combination the face verification performance (in terms of EER) and the attribute prediction performance from different function creep estimators are shown.
Figure \ref{fig:EER_Attr_estimation_b3_ColorFeret} and \ref{fig:EER_Attr_estimation_b4_ColorFeret} show the results for $k=3,4$ on ColorFeret.
Figure \ref{fig:EER_Attr_estimation_b3_Adience} and \ref{fig:EER_Attr_estimation_b4_Adience} show the same on Adience.
In these Figures the number of bins $k$ and the embedding sizes $L$ are analysed in the ranges of $k=[3, 4]$ and $L=[64, 4096]$.
All these scenarios show that a bigger embedding size $L$ leads to a lower face verification error. 
This observation agrees with the nature of positive-negative template comparisons.
Higher dimensional templates reduces the effect of random collisions for the positive-negative template comparison.
In lower dimensions, a random collision for an imposter comparison has a high impact on the resulting comparison score.
Towards the bin sizes $k$, it can be observed that $k=3$ has a lower face verification error than $k=4$, but also higher prediction accuracies from all function creep estimators.
Higher $k$ leads to more variabilities, which affects verification as well as the estimation of privacy-sensitive attributes.
These observations hold for both datasets and all function creep estimators.
Consequently, parameter $k$ and $L$ have to be chosen to accomplish the desired trade-off between attribute suppression and verification performance.

\subsection{Theoretical Reasoning Analysis}

\begin{figure*}[h]
\centering
  \subfloat[ColorFeret \label{fig:TheoryDistribution_ColorFeret}]{%
       \includegraphics[width=0.49\textwidth]{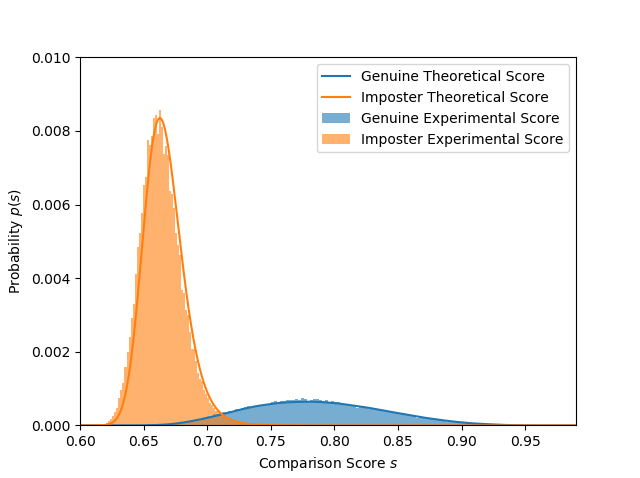}} \hspace{1mm}
  \subfloat[Adience \label{fig:TheoryDistribution_Adience}]{%
       \includegraphics[width=0.49\textwidth]{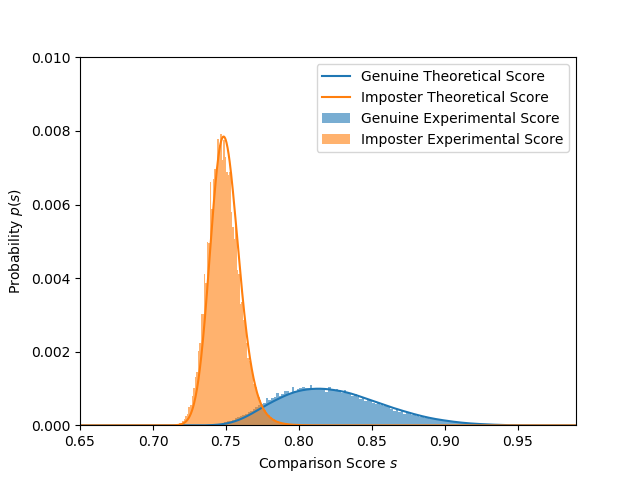}}
\caption{Validation of the theoretical reasoning: score distribution of the empirical data versus the theoretical predictions. The distributions show the genuine and imposter scores for $k=3$ bins. The theoretical score predictions are done with Equation \ref{eq:Proba_PosNeg} on the positive score distributions. The theoretical predictions accurately matches the experimental scores.}
\label{fig:TheoryDistributions}
\end{figure*}

In Section \ref{sec:Theory}, a theoretical reasoning for our negative face recognition approach was developed.
Here, we want to proof its correctness by empirically predicting the score distributions of our approach and comparing it with the achieved scores distributions.
For each comparison score in the positive domain, Equation \ref{eq:Proba_PosNeg} is used to calculate the most probable score in the negative-positive domain.
Repeating this process with every score in the distribution results in the score distributions in
Figure \ref{fig:TheoryDistributions}.
This figure shows the genuine and imposter scores distributions of our proposed approached with $k=3$, as well as its theoretically predicted distribution. 
It can be seen that on both databases, the predicted distributions accurately correspond to the empirical score distributions.
This validates our theoretical considerations from Section \ref{sec:Theory}.

\section{Conclusion}
Face biometric systems extract and store face templates during enrolment to enable the recognition of individuals during deployment.
However, privacy-sensitive information can be obtained from these templates.
Since many applications are expected to be used for recognition purposes only, this raises major privacy issues.
Previous work proposed supervised privacy-enhancing solutions that require training data with privacy-sensitive annotations, and thus, will only be able to suppress the attributes included in the training.
Moreover, these supervised solutions limit their application towards the suppression of a single attribute, increasing the risk towards function creep attacks on unconsidered attributes.
In this work, we successfully proposed negative face recognition, a privacy-enhancing solution working on the template-level.
It to prevents function creep attackers from successfully predicting privacy-sensitive information from stored face templates.
Our novel solution is based on the comparison of positive probe templates with negative reference templates.
While positive templates contain the facial properties of an individual, negative templates contain random complementary information, i.e. properties that the face do not have.
Since only negative templates are stored in the database, a reliable function creep estimation of privacy-sensitive information is prevented.
To guarantee a certain recognition performance, we further provided a theoretical foundation of our solution and proved its correctness empirically.
The experiments were conducted on two publicly available databases and on three privacy-sensitive attributes.
In the experiments, we simulated function creep attackers that know about the systems privacy mechanism and adapt their attacks based on it.
The experiments demonstrated the effectiveness of our approach under both, controlled and uncontrolled image capturing conditions.
Our proposed unsupervised solution significantly outperforms comparable approaches from previous work, while maintaining a significantly higher recognition performance.
In the uncontrolled scenario, negative face recognition fully retains the recognition performance while achieving suppression rates of up to 36\%.
Our solution is characterized by the fact that it prevents the accumulation of privacy-sensitive information during the training and offers more comprehensive privacy-protection.
Unlike previous work, negative face recognition is not limited to the suppression of single attributes.

Future work may investigate the proposed solution for the task of template protection, since our solution intrinsically provides its key properties: noninvertability, revocability, and nonlinkability.

%
%

\ifCLASSOPTIONcaptionsoff
  \newpage
\fi



%
{\small
\bibliographystyle{IEEEtran}
\bibliography{egbib}
}

%

%

\begin{IEEEbiography}[{\includegraphics[width=1in,height=1.25in,clip,keepaspectratio]{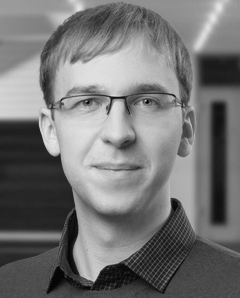}}]{Philipp Terh\"{o}rst} completed his studies in physics at the Technical University of Darmstadt in 2017. 
Since then, he has been working in the Smart Living \& Biometric Technologies department at the Fraunhofer Institute for Computer Graphics Research (IGD) as a research assistant. 
In the context of his doctorate, his field of work includes research into biometric solutions based on machine learning algorithms, specialising in dealing with reliability, privacy, and bias.
Furthermore, he is involved in the Software Campus program, a management program of the Federal Ministry of Education and Research (BMBF). He has served as a reviewer for various conferences and journals (e.g. TPAMI, IEEE Access, BTAS, ICB, FUSION).
\end{IEEEbiography}

\begin{IEEEbiography}[{\includegraphics[width=1in,height=1.25in,clip,keepaspectratio]{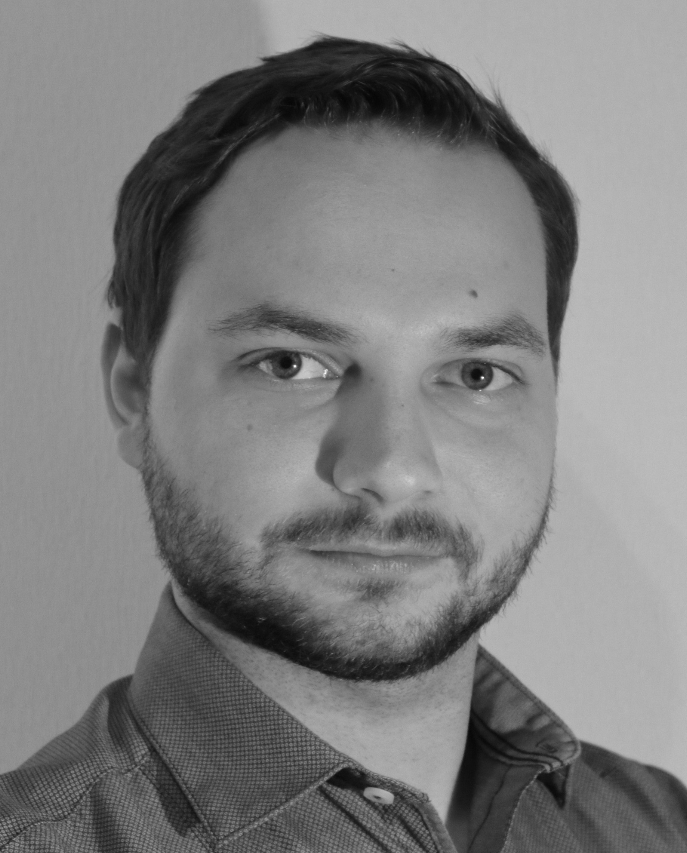}}]{Marco Huber} received his B.Sc. degree in computer science from the Technical University of Darmstadt in 2018.
He is currently working towards its M.Sc. degree in the same field. Moreover, he is working at Fraunhofer IGD since 2019. 
\end{IEEEbiography}

\begin{IEEEbiography}[{\includegraphics[width=1in,height=1.25in,clip,keepaspectratio]{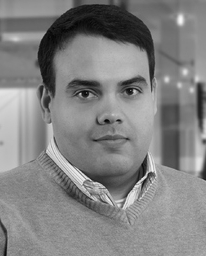}}]{Naser Damer} is a senior researcher at the competence center Smart Living \& Biometric Technologies, Fraunhofer IGD. He received his master of science degree in electrical engineering from the Technical University of Kaiserslautern (2010) and his PhD in computer science from the Technical University of Darmstadt (2018). He is a researcher at Fraunhofer IGD since 2011 performing applied research, scientific consulting, and system evaluation. His main research interests lie in the fields of biometrics, machine learning and information fusion. He published more than 50 scientific papers in these fields. Dr. Damer is a Principal Investigator at the National Research Center for Applied Cybersecurity ATHENE in Darmstadt, Germany. He serves as a reviewer for a number of journals and conferences and as an associate editor for the Visual Computer journal. He represents the German Institute for Standardization (DIN) in ISO/IEC SC37 biometrics standardization committee.
\end{IEEEbiography}

\begin{IEEEbiography}[{\includegraphics[width=1in,height=1.25in,clip,keepaspectratio]{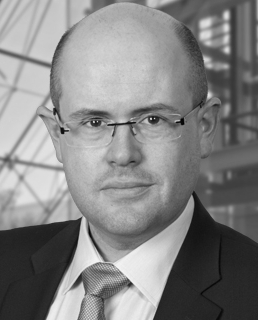}}]{Florian Kirchbuchner} is trained as an information and telecommunication systems technician and served as IT expert for the German Army from 2001 to 2009. Afterwards he studied computer science at Technical University of Darmstadt and graduated with a Master of Science degree in 2014. He has been working at Fraunhofer IGD since 2014, most recently as head of the department for Smart Living \& Biometric Technologies. He is also Principal Investigator at the National Research Center for Applied Cybersecurity ATHENE.
Mr. Kirchbuchner participated at Software Campus, a management program of the Federal Ministry of Education and Research (BMBF) and is currently doing his PhD at Technical University of Darmstadt on the topic "Electric Field Sensing for Smart Support Systems: Applications and Implications".
\end{IEEEbiography}

\begin{IEEEbiography}[{\includegraphics[width=1in,height=1.25in,clip,keepaspectratio]{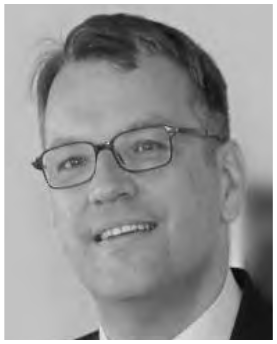}}]{Arjan Kuijper} received the M.Sc. degree in applied mathematics from Twente University, The Netherlands, the Ph.D. degree from UtrechtUniversity, The Netherlands, and the Habitation degree from TU Graz, Austria. He was an Assistant Research Professor with the IT University of Copenhagen, Denmark, and a Senior Researcher with RICAM, Linz, Austria. His research interests include all aspects of mathematics-based methods for computer vision, graphics, imaging, pattern recognition, interaction, and visualization. He is a member of the management of Fraunhofer IGD, where he is responsible for scientific dissemination. He holds the Chair in mathematical and applied visual computing with TU Darmstadt. He is the author of over 300 peer-reviewed publications,the Associate Editor of CVIU, PR, and TVCJ, the Secretary of the International Association for Pattern Recognition (IAPR), and serves both as a Reviewer for many journals and conferences, and as a program committee member and organizer of conferences.
\end{IEEEbiography}






\end{document}